# Agentic AI for Self-Driving Laboratories in Soft Matter: Taxonomy, Benchmarks, and Open Challenges

**Xuanzhou Chen**[1*], **Audrey Wang**[2], **Stanley Yin**[3], **Hanyang Jiang**[4], **Dong Zhang**[5*]
[1]School of Electrical and Computer Engineering, [4]School of Industrial and Systems Engineering, Georgia Institute of Technology. [2]Rice University. [3]Carnegie Mellon University. [5]Power Dream America, Inc.
[*] xchen920@gatech.edu and zhanglouis94@outlook.com

## Abstract

Self-driving laboratories (SDLs) close the loop between experiment design, automated execution, and data-driven decision making, and they provide a demanding testbed for agentic AI under expensive actions, noisy and delayed feedback, strict feasibility and safety constraints, and nonstationarity. This survey uses soft matter as a representative setting but focuses on the AI questions that arise in real laboratories. We frame SDL autonomy as an agent environment interaction problem with explicit observations, actions, costs, and constraints, and we use this formulation to connect common SDL pipelines to established AI principles. We review the main method families that enable closed loop experimentation, including Bayesian optimization and active learning for sample efficient experiment selection, planning and reinforcement learning for long horizon protocol optimization, and tool using agents that orchestrate heterogeneous instruments and software. We emphasize verifiable and provenance aware policies that support debugging, reproducibility, and safe operation. We then propose a capability driven taxonomy that organizes systems by decision horizon, uncertainty modeling, action parameterization, constraint handling, failure recovery, and human involvement. To enable meaningful comparison, we synthesize benchmark task templates and evaluation metrics that prioritize cost aware performance, robustness to drift, constraint violation behavior, and reproducibility. Finally, we distill lessons from deployed SDLs and outline open challenges in multimodal representation, calibrated uncertainty, safe exploration, and shared benchmark infrastructure.

## 1 Introduction

Self-driving laboratories (SDLs) are emerging as a practical paradigm for accelerating discovery in chemistry and materials, and they also provide a concrete, consequential setting for AI research [Abolhasani and Kumacheva, 2023]. In an SDL, an agent repeatedly selects the next experiment, executes it in the physical world, interprets measurements, and updates subsequent decisions under limited budgets of time, materials, and instrument availability [Bayley, *et al.*, 2025; Leong, *et al.*, 2025]. This loop instantiates sequential decision-making with expensive actions, noisy observations, and delayed feedback.

A central motivation for SDLs is the combinatorial explosion in modern experimental design. Even modest increases in the number of controllable steps can expand the candidate space from manageable to astronomically large, which makes exhaustive screening infeasible in both time and resource consumption (Fig. 1A) [Volk, *et al.*, 2023]. This scaling pressure means that autonomy is often necessary rather than optional, because progress depends on choosing informative experiments instead of enumerating the space. SDLs also surface challenges that are frequently understated in simulation-centric benchmarks. The environment is rarely stationary. Instruments drift, reagents vary across lots, and protocols introduce systematic effects. Agents must optimize scientific objectives while respecting feasibility and safety constraints, quantifying uncertainty, and handling operational realities such as batching, parallel experimentation, and occasional failures. These characteristics make SDLs a valuable testbed for agentic methods that must remain reliable when acting in the loop with the physical world [Zhang, *et al.*, 2024; Cornelisse, *et al.*, 2024].

This survey focuses on decision-making agents for SDLs in design–make–measure loops, with an emphasis on how agents choose experiments or protocol steps to achieve goals efficiently and reproducibly. We cover algorithmic approaches including Bayesian optimization (BO) and active learning for expensive evaluations [Liu, *et al.*, 2025], as well as reinforcement learning (RL) and planning methods for multi-step laboratory protocols. We also discuss tool-using or LLM-enabled orchestrators, but only when they correspond to verifiable decision policies. In this survey, verifiability means that proposed actions are grounded in structured specifications, executed through tools, and recorded in a form that can be audited and repeated (Fig. 1B). It separates the decision and control components that produce structured experiment specifications from the automated laboratory that executes them and returns multimodal observations and failure signals. It also includes optional human input for reward shaping and constraint editing.

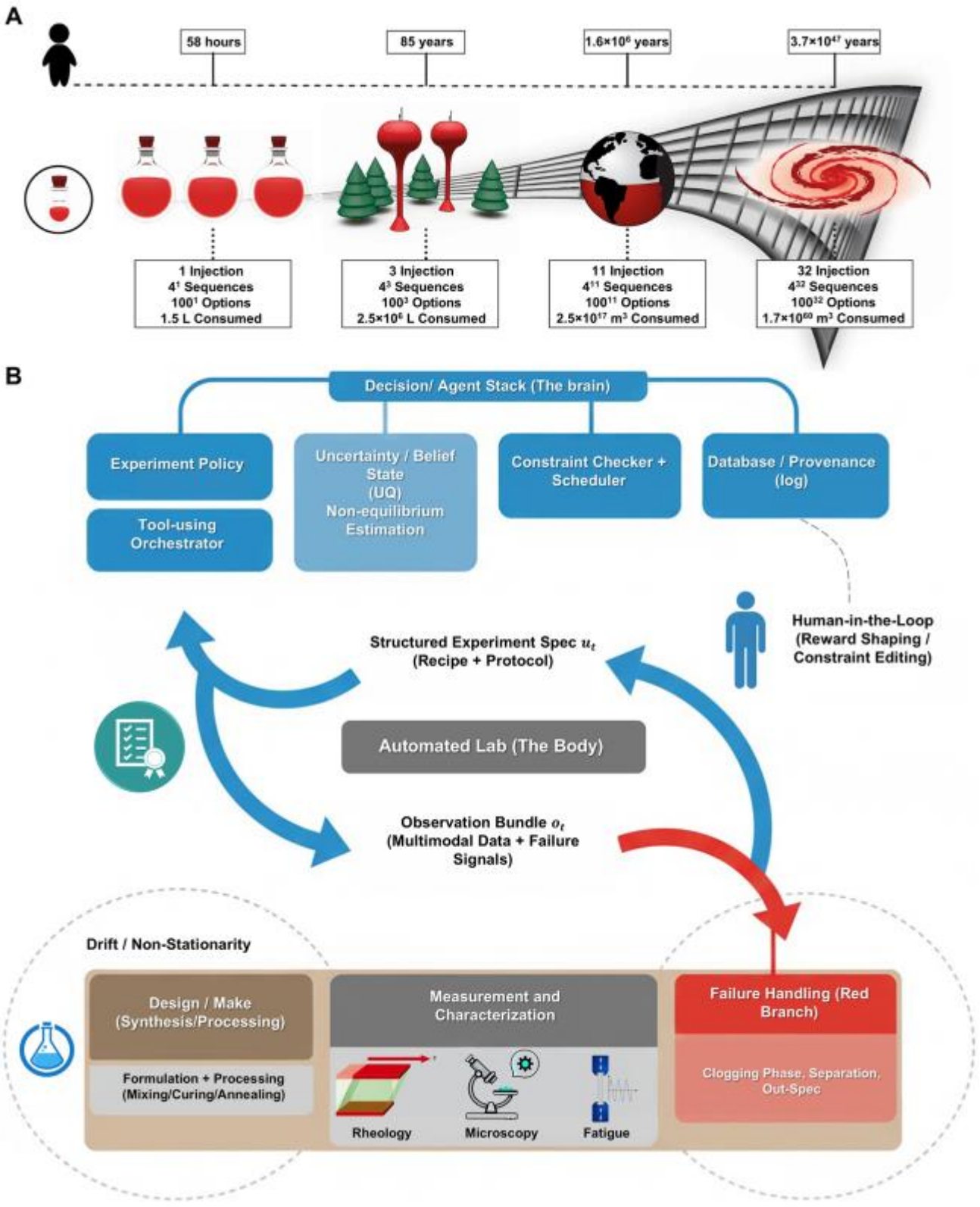

Figure 1: (A) Combinatorial explosion motivates automated experimentation. As the number of injections/decision points increases, the design space grows exponentially (*e.g.*, $4^n$ sequences and $100^n$ options), driving brute-force screening from hours (1 injection) to effectively impossible timescales (up to $\sim10^{47}$ years at 32 injections) and from liter-scale reagent use to astronomically large material requirements (up to $\sim10^{60}$ $m^3$). This highlights why scalable, automated/closed-loop experimentation is necessary to navigate high-dimensional search spaces. Reproduced with permission from Ref. [Volk, *et al.*, 2023]. Copyright 2023 Nature Publishing Group. (B) Overview of an agentic self-driving soft-matter lab, integrating automated hardware with decision-making agents.

Soft matter provides a particularly demanding context for this synthesis [Zhang, *et al.*, 2025; Chen, *et al.*, 2025]. Many targets depend on processing history, measurements are often multimodal, and useful protocols can span multiple stages. These properties amplify partial observability and long-horizon planning challenges, and they increase the importance of robustness under drift and operational constraints. Our contributions are fourfold. First, we formulate self-driving soft-matter laboratories as agent–environment interaction problems with explicit observations, costs, constraints, and delayed feedback, clarifying the decision structure behind experimental autonomy. Second, we provide a structured landscape that compares representative systems and distills recurring design patterns and bottlenecks across platforms. Third, we propose a concise taxonomy and a set of practical design principles that connect algorithmic choices, such as uncertainty modeling, constraint handling, decision horizon, and action parameterization, to system-level behavior and robustness. Finally, we outline evaluation protocols and benchmark tasks that emphasize sample efficiency, resource-aware performance, robustness to drift, and constraint-violation behavior, with the goal of making progress measurable and comparable.

## 2 Framing: experiments as agentic decision processes

### 2.1 Formalization

We model a self-driving laboratory as an agent interacting with an experimental environment over repeated decision cycles. At each iteration $t$, the agent maintains a history $h_t$ consisting of past experimental actions and observations, and selects the next experimental action $at$ from a feasible set $A(h_t)$. The laboratory then executes the experiment and returns an observation $o_t$, typically a collection of measurements such as scalar properties, spectra, rheology curves, or images, together with metadata about execution conditions and potential failure signals. The agent uses these observations to update its internal belief about the system and to decide what to do next [Häse, *et al.*, 2019; Bergamini, *et al.*, 2021].

This setup can be instantiated at different levels of temporal structure. In many SDL deployments, each experiment is treated as a single expensive query that yields an outcome $yt$ (possibly with uncertainty), and the decision problem is well approximated as a contextual bandit [Sekhari, *et al.*, 2023]. The context may include current constraints, available resources, or partial descriptors of previously observed structure–property relationships, and the goal is to select $at$ to maximize expected utility. When the experimental action itself is a sequence of steps, such as a multi-stage synthesis, processing route, or characterization protocol, the problem becomes a sequential decision process. In that case, the underlying dynamics can be described as an MDP when the state is sufficiently captured by the agent, or more realistically as a POMDP when only partial observations are available and relevant latent variables evolve over time. The distinction matters because it determines whether myopic acquisition is adequate or whether planning over multiple steps is needed to account for delayed effects and compounding uncertainty.

Constraints are central in physical experimentation and should be treated as first-class elements of the decision problem [Pooley and Wallace, 2022]. Feasibility constraints arise from chemistry and process limits (*e.g.*, solubility, viscosity ranges that a pipetting system can handle, or instrument operating windows). Safety constraints and compliance rules may prohibit certain compositions or procedures. Budget constraints include limited material, limited time, and instrument scheduling constraints, which often introduce an implicit cost $c(a_t)$ that varies across actions. A practical SDL agent therefore optimizes a utility function that trades off expected scientific value against costs, while controlling the probability or frequency of constraint violations. Many objectives are inherently multi-criteria [Rădulescu, *et al.*, 2020]. For example, an agent may seek high performance together with robustness, manufacturability, or stability under repeated cycling. This can be expressed through scalarization of multiple objectives into a single utility, through constrained formulations that treat some criteria as hard requirements, or through Pareto-style exploration where the goal is to learn and optimize trade-offs rather than a single optimum.

## 2.2 Soft matter specifics as AI challenges

Soft matter intensifies the challenges above because the mapping from actions to outcomes is frequently shaped by latent variables and hidden history. Processing history can determine microstructure and thus downstream properties, which means that the same nominal formulation can lead to different results depending on sequence, timing, and environmental conditions. From a decision-process viewpoint, this corresponds to non-Markov behavior under naive state definitions and motivates latent-state modeling or belief-based policies. In addition, soft-matter experimentation often yields observations that are not naturally summarized by a single scalar. Rheological sweeps produce curves, microscopy yields images, and stimuli-response tests produce time series with hysteresis and fatigue effects. The agent must therefore operate on multi-modal observations and learn representations that preserve uncertainty, rather than relying on clean, low-dimensional feedback.

Many soft-matter goals require long-horizon protocols, which includes multi-stage synthesis, curing, annealing, or sequential processing steps. These steps introduce delayed and sometimes irreversible effects. Characterization is sequential, for example it uses low-cost tests to screen candidates, then high-cost measurements to confirm performance. These settings push the SDL away from one-step optimization toward planning problems where short-term information gain must be balanced against long-term objective improvement. Nonstationarity is also common in designing soft matter. As instruments drift, reagents age, and ambient conditions change, an agent with a fixed objective surface can degrade over time. This motivates strategies for drift detection, continual adaptation, and robust decision-making under distribution shift.

Finally, operational factors are not merely engineering details; they shape the decision process. Experiments are frequently executed in batches for throughput, which couples decisions across candidates and introduces exploration design questions under parallelism. Failures are unavoidable and can be informative. An agent must decide whether to retry, adjust a protocol, or avoid regions of the action space that trigger systematic failure. The combination of latent history, multi-modal observations, long-horizon protocols, nonstationarity, and batching creates a decision environment where the central AI problem is not only finding high-performing candidates, but doing so with reliability, safety, and reproducibility under real-world constraints.

## 2.3 Minimal primer of agent families for SDLs

Bayesian optimization (BO) and active learning are natural starting points for SDLs because they are designed for settings where each experiment is costly and the number of trials is limited [Fiore, *et al.*, 2023]. In a typical formulation, the agent builds a probabilistic surrogate model that maps the experimental action $a$ to an outcome $y$, together with an estimate of uncertainty. Decisions are made by optimizing an acquisition function that trades off exploitation and exploration, for example by favoring actions that are predicted to perform well while still reducing uncertainty in promising regions [Ndikum and Ndikum, 2024]. This family is particularly well suited to one-step or weakly sequential settings where each trial can be treated as an independent query, and where the key difficulty lies in sample efficiency under noise. In practice, SDLs often require extensions beyond the textbook formulation, including constraints that separate feasible from infeasible regions, heteroscedastic or non-Gaussian noise, and batch selection when multiple experiments are run in parallel (Fig. 2A). A useful mental model for this survey is that Bayesian optimization and active learning offer a principled mechanism for short-horizon experiment selection with uncertainty quantification, and they often serve as strong baselines even when richer sequential structure exists [Hvarfner, *et al.*, 2023].

Reinforcement learning (RL) and planning methods become important when the experimental action is itself a sequence of interdependent steps and when outcomes depend on latent history [Eysenbach, *et al.*, 2020]. In these settings, the agent is no longer choosing a single composition or parameter vector, but rather a protocol that may include branching decisions, intermediate measurements, retries, and adaptive stopping. The reward may be delayed until the end of a protocol, and intermediate observations may be used to update beliefs and steer subsequent steps. Planning-based views are especially relevant when the protocol structure is known and can be parameterized, since the agent can search over structured action spaces and incorporate constraints explicitly. RL views are especially relevant when dynamics or failure modes are complex and must be learned from data, or when the agent must trade off exploration, exploitation, and risk over multiple steps [Cao, *et al.*, 2025]. For SDLs, the practical bottleneck is often data efficiency and safety. Pure online RL is rarely affordable in physical labs, which motivates approaches that exploit prior knowledge, simulators, offline logs, and conservative learning objectives. For the rest of the paper, it is helpful to treat RL and planning as the primary tools for long-horizon protocol optimization, failure-aware decision-making, and settings where myopic acquisition strategies are insufficient (Fig. 2B).

LLM-enabled, tool-using orchestrators have recently been proposed as a way to broaden SDL autonomy beyond parameter optimization, by generating experiment plans, selecting tools, querying past results, and coordinating multiple subsystems [Hong, *et al.*, 2025]. For this survey, the key question is not whether an LLM can produce plausible text, but whether it can implement a decision policy that is auditable, reproducible, and grounded in the SDL's available interfaces. A useful lens is to view an LLM as a high-level controller that proposes candidate actions or plans, while verification and execution are delegated to structured tools such as constraint checkers, planners, database queries, and experiment controllers. Under this lens, provenance and validation become central. The system should record what information was used, what tools were called, what constraints were checked, and what experiment specification was ultimately executed. This perspective also clarifies the limits of the approach. Without grounding and verification, an LLM's suggestions can be difficult to trust, hard to reproduce, and prone to subtle violations of feasibility or safety constraints [Pantha, *et al.*, 2024]. When paired with tool-based checking and explicit logging, however, LLM orchestration can complement algorithmic agents by improving interface flexibility, supporting human-in-the-loop steering, and handling heterogeneous tasks such as documentation, error triage, and protocol templating.

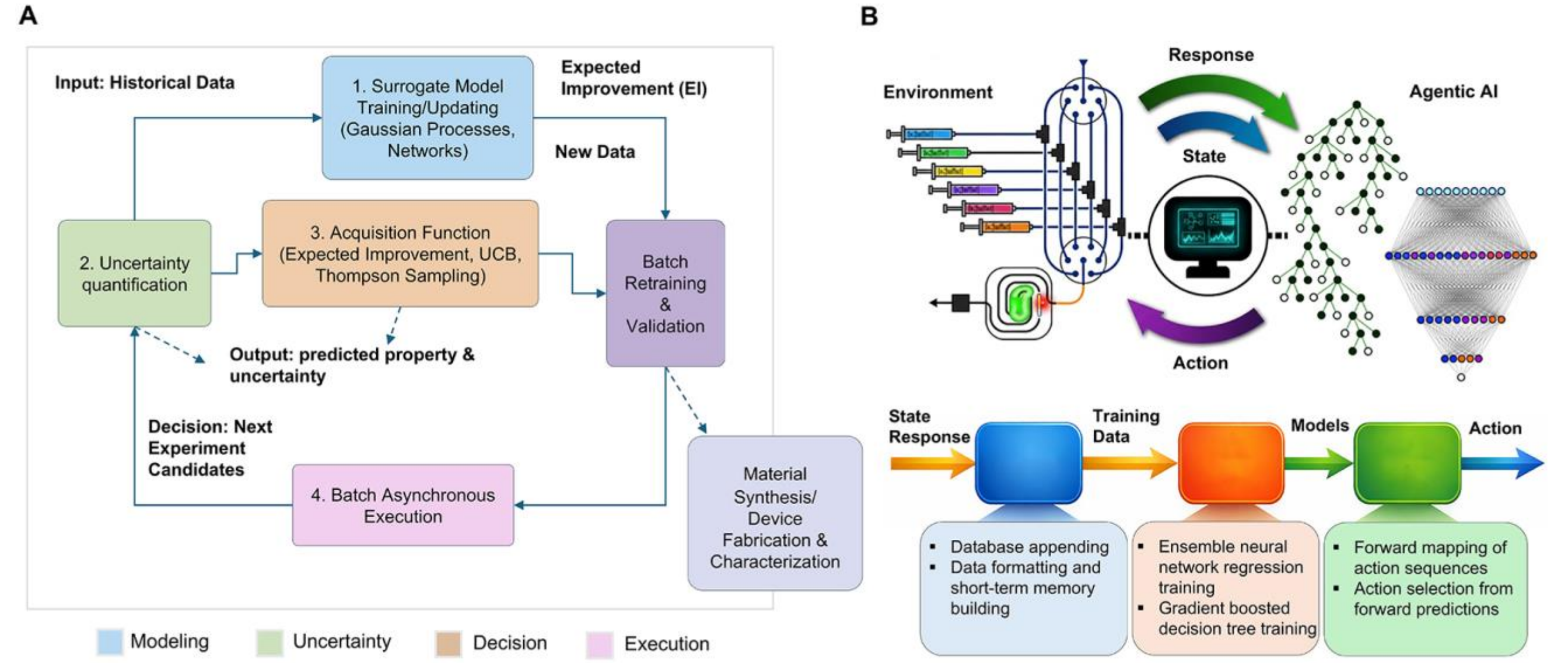

Figure 2: (A) Bayesian optimization and active-learning workflow for self-driving laboratories, including surrogate-model training, uncertainty quantification, acquisition-driven candidate selection, batch execution, and retraining/validation. (B) RL feedback loop between the automated experimental platform and the learning agent, illustrating how state–action–response signals guide model updates and subsequent condition selection. Reproduced with permission from Ref. [Volk, *et al.*, 2023]. Copyright 2023 Nature Publishing Group.

# 3 Structured landscapes

A large fraction of the self-driving laboratory literature is presented as platform demonstrations or application case studies, which makes direct comparison difficult and often blurs the underlying AI contributions. For an AI-for-science survey, the key is to abstract away from lab-specific idiosyncrasies while keeping enough scientific detail to make the decision problem concrete. In soft-matter SDLs, the agent's action space is typically a structured combination of formulation and processing. Formulation decisions specify compositions on a simplex, such as monomer and crosslinker ratios in hydrogels, polymer blend fractions, solvent or salt concentrations, or additive and nanoparticle loadings [Timaite and Castelle 2025]. Processing decisions include mixing order and speed, temperature profiles, curing or annealing schedules, shear or flow conditions, and aging time, all of which can reshape microstructure and therefore final properties. Observations are rarely a single scalar. They often include rheology curves, full stress–strain responses, swelling or phase-transition measurements, microscopy images of morphology, scattering profiles, and time series for stimuli-responsive actuation and fatigue. These characteristics make soft matter an especially transparent instance where an autonomous agent must operate under feasibility constraints, history dependence, multi-modal feedback, and nonstationarity.

To revisit prior work, we summarize representative systems and decision-problem archetypes that emphasizes decision-relevant structure rather than narrative descriptions [Alqithami, 2025]. Table 1 includes the platform and domain, the action space, the observation modality, the objective or utility definition, the mechanism for constraint and failure handling, the decision method, and a key takeaway from an agent-design perspective. To avoid conflating evidence with aspiration, we also indicate the implementation status of each row (demonstrated, partial, or proposed). This representation enables a capability-driven comparison: it highlights which decision problems are solved end-to-end in practice, which are only partially realized, and where laboratory realities—mixed-variable actions, batching and parallel execution, failure modes, and drift—remain under-modeled. We organize the landscape into four capability buckets; while mature SDLs may span multiple buckets, each bucket captures a dominant challenge that shapes agent design.

*Formulation-only optimization is the most common entry point*: each action proposes a formulation (and occasionally a few process knobs), and each trial returns a scalar or low-dimensional outcome (*e.g.*, modulus, swelling, transition temperature, or a proxy score). Typical cases include hydrogel formulation over monomer/crosslinker/initiator, polymer blends over component fractions, and colloid/surfactant searches over salt and additive loading. Decision-wise, this setting is well captured by expensive black-box optimization, so Bayesian optimization and active learning dominate. The hard parts are usually feasibility and feedback quality: simplex and solubility constraints, viscosity/processability limits for automation, phase separation or gelation that blocks measurement, and heteroscedastic noise that depends on protocol choices. Because throughput is often parallel, batch selection and diversity are part of the decision problem, not an afterthought.

*Process and protocol optimization is required when outcomes depend on trajectories rather than final composition*. Soft-matter protocols often involve order-sensitive mixing/dissolution, temperature ramps, curing schedules, shear alignment, staged aging, and multi-stage characterization. This introduces delayed effects, partial observability, and mid-protocol failures (*e.g.*, clogging/gelation, incomplete curing, out-of-range states). While RL and planning provide useful lenses, practical systems favor structured protocol parameterizations, safety constraints, and conservative learning from scarce trials and offline logs. A recurring theme is failure-aware autonomy: treating execution failures as informative and explicitly learning feasibility boundaries can substantially reduce wasted runs. Resource coupling (shared equipment/time) further makes scheduling and batching constraints first-class.

*Closed-loop characterization and representation learning shifts the bottleneck to measurement*: agents must turn rich, noisy observations (rheology sweeps, stress–strain curves, microscopy images, scattering profiles, and time-series signals) into decision-relevant variables with calibrated uncertainty. Approaches range from physics-motivated features to learned embeddings and predictive models; in either case, calibration matters more than raw accuracy under drift, protocol changes, and distribution shift. This bucket also yields cost-aware sequential decisions: choosing which measurement to run, when to stop, and how to allocate budget across low-cost screening and expensive confirmatory tests, naturally motivating multi-fidelity policies and utilities based on time, cost, and information gain.

*Human-in-the-loop and tool-using agents address ambiguity and operational complexity*. Soft-matter experiments frequently yield borderline or partially failed outcomes that still carry signal, requiring human judgment to refine objectives, labels, and constraints. Tool-using agents (including LLM-enabled orchestrators) can improve autonomy by translating high-level goals into structured experiment specifications, retrieving prior results, coordinating subsystems, and supporting failure recovery. For AI-for-science, the key differentiator is whether the system implements verifiable decision policies: constraint checking, planning, explicit logging, and auditable, reproducible action representations matter as much as suggestion quality.

Cross-cutting lessons: *i*) Specification dominates: action parameterization, constraint encoding, and utility definitions often matter more than small algorithmic choices. *ii*) Constraints and failures are signals; modeling infeasibility/execution failure improves robustness and efficiency. *iii*) Parallelism changes the problem: batch selection, scheduling, and cost-aware utilities are essential. *iv*) History dependence motivates latent-state reasoning beyond myopic selection. *v*) Representation learning must preserve uncertainty; uncalibrated models lead to brittle closed-loop behavior. *vi*) Nonstationarity is the default, so drift detection and adaptation are required. *vii*) Tool-using autonomy requires provenance and verifiability: structured actions and auditable records are prerequisites for reliable deployment

# 4 Taxonomy and design principles

Agentic self-driving laboratories can look different in implementation, but most variation can be explained by a small set of design choices about decision structure and interface rather than specific algorithms [Wu, *et al.*, 2023]. A practical taxonomy should therefore describe how far decisions must remain coherent across time, what the agent can directly control, and how the loop stays safe and reliable under imperfect execution. *A first discriminator is the decision horizon.* Some loops are well approximated as selecting the next experiment as an expensive query, while others require coherent reasoning over interdependent sequences because protocol choices, intermediate measurements, and delayed effects determine outcomes. *The second discriminator is the action interface.* Real SDL action spaces are mixed and constrained, combining continuous variables with discrete choices such as reagents, templates, instruments, and measurement selections. Expressing actions as structured objects such as parameterized protocols allows feasibility checks, safety rules, and scheduling constraints to be enforced directly and improves generalization across protocol families.

Uncertainty and feedback modeling is another core axis. Physical experimentation exhibits heteroscedastic and protocol dependent noise, censored outcomes from failures, and nonstationarity due to drift [Villalobos-Arias, *et al.*, 2023]. When observations are rich, uncertainty must remain calibrated after representation learning, otherwise closed loop decisions become overconfident and fragile. Constraints and failure handling are not side conditions: feasibility, safety, budget, and batching define what science is reachable.

Systems differ in whether constraints are enforced by filtering, explicit constrained optimization, or learned feasibility models, and in whether failures are treated as missing data or logged as structured signals that improve future decisions. Human involvement and embodiment complete the picture. Many deployments require oversight for objective definition, constraint updates, batch approval, and recovery from cascaded failures. As coupling to execution increases, operational decisions such as batching, measurement selection, and resource allocation become inseparable from scientific ones. These axes motivate recurring principles: encode feasibility in action parameterizations to reduce wasted trials, maintain calibrated uncertainty under drift to support safe exploration, log failures with interpretable causes, and implement provenance and structured specifications so decisions can be audited and rerun, especially for tool using and LLM enabled components.

| # | Platform | Inputs & Controls | Objective & Constraints | Decision method | References |
|---|---|---|---|---|---|
| **1** | Hydrogel formulation SDL | Composition on simplex; Scalar property + failures | Target properties; Solubility/viscosity limits | Constrained / batch BO | Partially demonstrated [Zaki, *et al.*, 2025; Rooney, *et al.*, 2022] |
| **2** | Polymer blend/ colloid search | Simplex fractions; $T_g$, mechanical proxies, features | Property matching; Phase separation/processability window | Simplex surrogate + BO/AL | Partially demonstrated [Vriza, *et al.*, 2023] |
| **3** | Order-sensitive mixing protocols | Step ordering/mixing speed; final outcome | Final property; failure-rate minimization; logging/gelation mid-protocol as labels | Latent-state modeling | Partially demonstrated [Yoo, *et al.*, 2025] |
| **4** | Aging/annealing optimization | Aging time, ramps, shear profiles; Rheology curves | Property/structure targets; Equipment/time coupling | Multi-stage policy / updates | Yes [MacLeod, *et al.*, 2020] |
| **5** | Rheology-loop SDL | Formulation/process, measurement selection; Rheology | Property target; Heteroscedastic noise; calibrated UQ | Functional surrogate/calibrated UQ | Yes [Beaucage, *et al.*, 2024] |
| **6** | Morphology optimization | Formulation/process, imaging; morphology features | Morphology target; Focus drift, imaging failure; QC | Embedding AL/ drift checks | Partially demonstrated [Li and Yi, 2025] |
| **7** | Scattering-guided structure control | Measurement, formulation/process; Scattering profiles/structures | Info gain / structure estimation accuracy; Budget/time | Bayesian ED/MF | Yes [Wu, *et al.*, 2025] |
| **8** | Budgeted multi-fidelity screening | Proxy vs confirmatory tests; Proxy/confirmatory outputs | Cost-aware utility; Early stop; Budget | Value-of-information policies | Partially demonstrated [Lo, *et al.*, 2023] |
| **9** | Human-assisted boundary-case interpretation | Agent proposes; human labels; Partial failures signal | Robust optimization; Structured logging | Mixed-initiative AL | Yes [Tubaro, *et al.*, 2020] |
| **10** | Tool-using orchestrator | Goals, experiment specs; Logs + DB + instruments | Task + reproducibility; Rules; provenance; handling | LLM/checker/planner | Early but real [Salimpour, *et al.*, 2025] |

Table 1: Representative systems are summarized by decision-relevant structure (*i.e.*, platform/domain, action space, observation modality, objective/utility, constraint and failure handling, decision method, and an agent-design takeaway). The final column indicates whether the capability is demonstrated end-to-end in the literature or remains only partially realized. Rows 1-2 correspond to formulation-only optimization, rows 3-4 to process/protocol optimization, rows 5-8 to closed-loop characterization and representation, and rows 9-10 to human-in-the-loop and tool-using agents.

# 5 Evaluation: benchmarks and metrics

Meaningful evaluation for self-driving laboratories should emphasize decision quality under real experimental constraints, rather than idealized performance in unconstrained and stationary settings. Methods should be judged by resource efficiency, reliability under feasibility and safety constraints, and robustness to noise, occasional failures, and drift. Benchmarks should therefore expose constrained and mixed action spaces, support batch or parallel execution when relevant, and remain reproducible and shareable [Hou, *et al.*, 2016]. A practical design is to start from offline logs and curated datasets that capture real variability, then augment them with simulators or semi-synthetic generators that preserve key statistical and physical properties while enabling controlled, scalable comparisons. We propose three benchmark task templates that capture common decision challenges while remaining broadly useful: *i*) LCST and transition targeting, where actions specify a formulation (e.g., on a compositional simplex) with a small set of processing parameters, observations include transition curves or derived assay features, and the goal is to reach a target switching threshold under feasibility constraints. *ii*) Mechanical multi-objective trade-offs, where actions propose candidate designs and processing choices, observations include full stress–strain curves and derived features, and the goal is to optimize competing objectives under limited evaluation budgets and practical constraints (*e.g.*, manufacturability or stability). *iii*) Stimuli-responsive actuation with degradation, where observations are time-series responses over repeated cycles, objectives include amplitude and kinetics as well as reversibility and fatigue resistance, and nonstationarity emerges through aging and drift.

Protocols should evaluate behavior throughout the loop, not only the final best material. Report sample efficiency as cost or time to reach a target utility or success probability, and report best-found performance under a fixed budget. Robustness can be measured by introducing controlled drift or distribution shift in the generator and quantifying performance degradation. Constraint violation rate directly captures safety and operational reliability, including responses after failures. Reproducibility should be an explicit checklist item, covering specification of action space, constraints, cost and noise models, budgets, seeds, and released logs or offline data. Baselines should be simple and transparent, including random and space-filling designs, standard BO for short-horizon tasks, and a constrained BO variant to reflect feasibility modeling [Wei, *et al.*, 2025]. For protocol-oriented tasks, include a lightweight planning or RL baseline, such as a template-based policy that searches over a parameterized protocol family using a simple model-based planner or conservative offline updates. The goal is to provide stable reference points that reflect decision structure, not to maximize baseline performance.

# 6 Lessons learned and research agenda

Across existing self-driving laboratory systems, autonomy fails less often because an optimizer is slightly suboptimal, and more often because the decision loop becomes unreliable when confronted with physical reality. The most frequent failures involve constraint violations and brittle feasibility modeling. When feasibility is not modeled explicitly, agents waste trials proposing non-executable actions such as phase-separating formulations, clogged dispensing, premature gelation, or instrument settings outside operating windows. Even with explicit constraints, overconfident models can repeatedly select actions that appear promising under a surrogate but fail under measurement noise, hidden protocol sensitivity, or unobserved confounders. Nonstationarity further amplifies these issues. Instrument drift, reagent aging, ambient variation, and latent state shifts can quickly degrade policies that assume stationarity. Tool-using and LLM-enabled components add another layer of risk: plausible textual plans are not executable protocols unless systems enforce structured outputs, constraint checks, and provenance logging. Overall, the dominant bottlenecks are robustness, safety, and reproducibility, not only sample efficiency.

These observations motivate several open research problems broadly relevant to agentic AI. *i*) Latent-state and belief modeling for partially observed experimental dynamics with history dependence. *ii*) Safe exploration under hard, time-varying constraints, where feasibility boundaries are complex and only partially observable through failures. *iii*) Multimodal representation learning with calibrated uncertainty, enabling reliable decisions from curves, images, spectra, and time series under distribution shift. *iv*) Failure-aware autonomy, including diagnosis, retry and protocol adaptation, and continual updates of feasibility models rather than skipping failed trials. *v*) Verifiable tool-using agents, where provenance, auditability, and rerun-ability are built into policy design through structured interfaces and explicit verification. Moreover, progress will accelerate with shared resources and reporting norms. Community datasets and offline logs should include failures, costs, and contextual metadata in addition to successful outcomes. Standardized reporting should specify action spaces and parameterizations, feasibility and safety constraints, cost and time accounting, batching assumptions, and noise and drift characteristics. Benchmarks and leaderboards can consolidate evaluation practice, but they should prioritize decision quality under constraints and reproducibility rather than single-number performance in idealized settings.

# 7 Conclusion

Self-driving laboratories provide a concrete and high-impact setting for agentic AI, where decisions must be made under expensive experimentation, noisy and delayed feedback, hard feasibility constraints, and nonstationarity. Using soft matter as a motivating domain, this survey framed SDLs as decision processes, synthesized the landscape through capability-driven buckets, and proposed a concise taxonomy that clarifies how horizon, action structure, uncertainty, constraints, human involvement, and embodiment shape real-world autonomy. We also outlined benchmark tasks and evaluation metrics that prioritize decision quality under realistic costs and constraints, and summarized lessons that highlight current failure modes and research opportunities. Taken together, the taxonomy, benchmarks, and research agenda aim to support more reliable, comparable, and reproducible progress on agentic AI systems that can operate robustly in scientific laboratories.

## Acknowledgment

We thank Michael Meram from Alti Company to provide support and assistance on AI agents.

## Ethical Statement

There are no ethical issues.